\documentclass{article}
\usepackage{spconf,amsmath,graphicx,hyperref}
\usepackage{booktabs}
\usepackage{tikz}
\usetikzlibrary{arrows.meta,positioning}
\definecolor{bridgeBlue}{HTML}{4E79A7}
\definecolor{bridgeTeal}{HTML}{76B7B2}
\definecolor{bridgeOrange}{HTML}{F28E2B}
\definecolor{bridgePurple}{HTML}{B07AA1}
\definecolor{bridgeInk}{HTML}{30343B}
\definecolor{bridgePaper}{HTML}{F5F7FA}

\newcommand{\method}{\textsc{BridgeMem}}

\title{BRIDGEMEM: CAUSAL DYADIC TRANSITION RESIDUALS FOR TEMPORAL KNOWLEDGE GRAPH FORECASTING}
\name{Zeyan Li$^1$, Libing Chen$^2$, Shengda Zhuo$^3$, Yin Tang$^3$, Jianfeng Xu$^1$\sthanks{Corresponding author.}}
\address{$^1$ Shanghai Jiao Tong University, $^2$ University of Chicago, $^3$ Jinan University}

\begin{document}
\ninept
\maketitle
\begin{abstract}
Temporal knowledge graph forecasting aims to infer future relational facts from the temporal structure of observed events. Existing forecasters mainly summarize history through entity states, relation states, paths, or exact recurrence. These views often miss pair-specific transition evidence, that is, the way prior relations between the query actor and a candidate change the odds of the target relation. We introduce \method{}, which estimates this quantity as a residual added to the log scores of a frozen full-vocabulary forecaster. For each candidate, \method{} retrieves the pair's events that strictly precede $t$, encodes their relations, directions, and lags, and converts them into a likelihood-ratio correction. A support-adaptive empirical-Bayes reader trusts exact transition counts where they are abundant and backs off to a learned attention estimator where they are sparse. The backbone's own uncertainty gates the correction, so confident queries and candidates without dyadic history are left unchanged. On five benchmarks, \method{} improves on the strongest of nine baselines from 2021--2026 in all 20 filtered MRR and Hits@$\{1,3,10\}$ comparisons, with MRR gains of 0.0213, 0.0164, 0.0216, 0.0112, and 0.0028 over the best prior result. These results show the value of explicit dyadic transition modeling.
\end{abstract}
\section{Introduction}

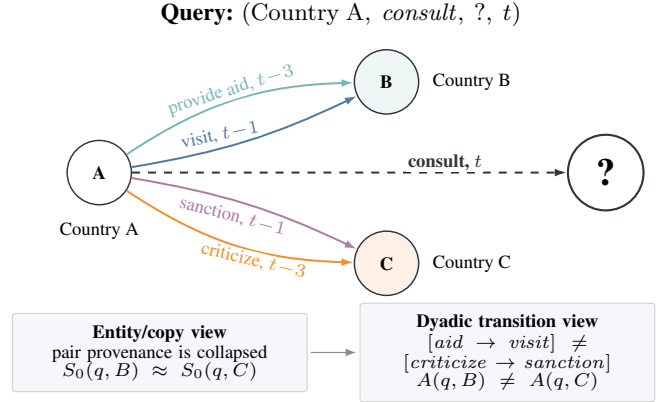
\begin{figure}[!t]
\centering
\begin{tikzpicture}[
  x=1mm,y=1mm,font=\scriptsize,
  actor/.style={circle,draw=bridgeInk,line width=.55pt,fill=white,
                minimum size=8.5mm,inner sep=0pt,font=\bfseries\scriptsize},
  event/.style={-{Latex[length=1.6mm,width=1.05mm]},line width=.75pt},
  offlabel/.style={fill=white,inner sep=.55pt,align=center},
  note/.style={draw=black!18,rounded corners=1.2pt,fill=bridgePaper,
               align=center,minimum height=8mm,text width=36mm,
               inner sep=1.5mm}
]
\node[font=\bfseries\small] at (44,49)
  {Query: $(\mathrm{Country\ A},\ \mathit{consult},\ ?,\ t)$};

\node[actor] (a) at (12,28) {A};
\node[actor,fill=bridgeTeal!12] (b) at (50,40) {B};
\node[actor,fill=bridgeOrange!12] (c) at (50,16) {C};
\node[circle,draw=bridgeInk,line width=.8pt,minimum size=10mm,
      font=\bfseries\Large] (unknown) at (79,28) {?};

\node[anchor=north] at (12,22.6) {Country A};
\node[anchor=west] at (55,40) {Country B};
\node[anchor=west] at (55,16) {Country C};

\draw[event,bridgeTeal,bend left=16] (a) to
  node[pos=.50,sloped,above,fill=white,inner sep=.5pt]
  {provide aid, $t\!-\!3$} (b);

\draw[event,bridgeBlue,bend right=8] (a) to (b);
\node[offlabel,rotate=12,text=bridgeBlue]
  at (28.0,33.0) {visit, $t\!-\!1$};

\draw[event,bridgeOrange,bend right=16] (a) to
  node[pos=.6,sloped,below,fill=white,inner sep=.45pt]
  {criticize, $t\!-\!3$} (c);

\draw[event,bridgePurple,bend left=8] (a) to (c);
\node[offlabel,rotate=-15,below,text=bridgePurple]
  at (30.0,23.5) {sanction, $t\!-\!1$};

\draw[event,bridgeInk,dashed] (a) --
  node[pos=.72,above,fill=white,inner sep=.6pt,font=\bfseries\scriptsize]
  {consult, $t$} (unknown);

\node[note] (collapsed) at (20,4.2) {\textbf{Entity/copy view}\\
  pair provenance is collapsed\\[-.3mm]
  $S_0(q,B)\approx S_0(q,C)$};

\node[note] (dyadic) at (66,4.2) {\textbf{Dyadic transition view}\\
  $[\mathit{aid}\!\rightarrow\!\mathit{visit}]\ne
   [\mathit{criticize}\!\rightarrow\!\mathit{sanction}]$\\[-.3mm]
  $A(q,B)\ne A(q,C)$};

\draw[-{Latex[length=1.4mm]},black!45,line width=.45pt]
  (40,4.2) -- (46,4.2);
\end{tikzpicture}
\caption{The resolution mismatch. Entity/copy views collapse the queried actor--candidate dyad, while typed dyadic histories preserve distinct transition evidence.}
\label{fig:motivation}
\end{figure}

Temporal knowledge graph (TKG) forecasting answers queries such as $(a,r,?,t)$ by ranking entities from the events observed before $t$. Although the evaluation is a ranking task, the evidence supporting a candidate is inherently dyadic. It depends on the target relation $r$ and on what has happened between the query actor $a$ and that specific candidate. For a future consultation between two countries, as in $(\mathrm{Country\ A}, \mathit{consult}, ?, t)$, earlier aid and visits argue for it, while earlier disputes argue against it (Figure~\ref{fig:motivation}). Existing forecasters, however, fail to capture this signal. Entity states summarize a candidate's neighborhood without recording which events involved the query actor, and copy distributions fire only when the exact same relation has occurred on the pair before. As a result, two candidates with similar entity-level states receive similar scores, even if only one has a direct interaction history with $a$. What is missing is the quantity both views discard, that is, how much the observed sequence of relations on the actor--candidate pair should shift the odds of the queried relation.

Existing TKG forecasters can be broadly grouped by how they read history. One family models the continuous evolution of facts. Know-Evolve uses a temporal point process \cite{trivedi2017knowevolve}, RE-Net recurrently aggregates event history \cite{jin2020renet}, RE-GCN evolves entity and relation states over snapshots \cite{li2021regcn}, TiRGN combines local and global recurrence \cite{li2022tirgn}, and EST tunes entity states to harmonize structure with sequence \cite{li2026est}. These methods capture temporal dynamics effectively, but they summarize a candidate through entity or relation states. The resulting evidence is not addressed to the ordered pair between the query actor and the candidate, so it cannot record which cross-relation events actually occurred on that pair.

A second family exploits recurrence and explicit reasoning structures. CyGNet copies repeated facts \cite{zhu2021cygnet}, CENET separates historical from non-historical candidates \cite{xu2023cenet}, and HRI retrieves historical repetitions \cite{gastinger2024hri}. CluSTeR, xERTE, TimeTraveler, TLogic, TPAR, CSI, CoH, and INFER go further by retrieving clue graphs, expanding query-relevant subgraphs, searching temporal paths, applying temporal logical rules, or separating causal subhistories \cite{li2021cluster,han2021xerte,sun2021timetraveler,liu2022tlogic,chen2024tpar,chen2024csi,xia2024coh,li2025infer}. Dynamic graph memories such as JODIE, DyRep, and TGN provide another variant, storing time-stamped node memories that evolve after interactions \cite{kumar2019jodie,trivedi2019dyrep,rossi2020tgn}. Despite their differences, these methods usually address history at a coarser resolution than the final ranking decision. They can model entity activity, relation recurrence, and temporal paths, but they do not assign a separate transition effect to the ordered pair $(a,c)$. This leaves no explicit dyadic quantity for the prior events on $(a,c)$, a term that preserves directionality, recency, and relation-specific transition evidence for target relation $r$.

Capturing it calls for a memory addressed by the ordered actor--candidate pair, whose stored value keeps the transition sequence intact. Such a memory must also distinguish a reliable repeated transition from a one-off sparse observation, since an exact lookup alone would merely exchange representation loss for frequency noise. We propose \method{}, which estimates this quantity as a candidate-specific likelihood-ratio residual on top of a frozen full-vocabulary forecaster. For each subject--candidate pair, \method{} retrieves the strictly prior events on the pair and forms two estimates of the next relation, an attention encoder over relation, direction, and recency tokens, and a smoothed count of the same transition contexts. A support-dependent mixture uses counts when evidence is abundant and the neural estimate when it is sparse. The evidence is expressed relative to the marginal target-relation prior, so common relations do not receive an indiscriminate boost. The frozen forecaster's own oracle uncertainty gates the residual, so confident decisions change little. Ambiguous queries, by contrast, admit a larger correction. Candidates without dyadic history keep their original scores. This work makes three contributions.
\begin{itemize}
  \item We formulate candidate-ranking evidence as a dyadic relation-transition likelihood ratio instead of another entity representation.
  \item We introduce a support-adaptive count/neural estimator and an oracle-uncertainty gate that improve a frozen forecaster without end-to-end retraining.
  \item We show on five benchmarks that \method{} improves over nine prior methods on all filtered MRR and Hits@$\{1,3,10\}$ comparisons.
\end{itemize}

\section{BridgeMem}

\begin{figure*}[!t]
\centering
\resizebox{\textwidth}{0.40\textwidth}{%
  \includegraphics{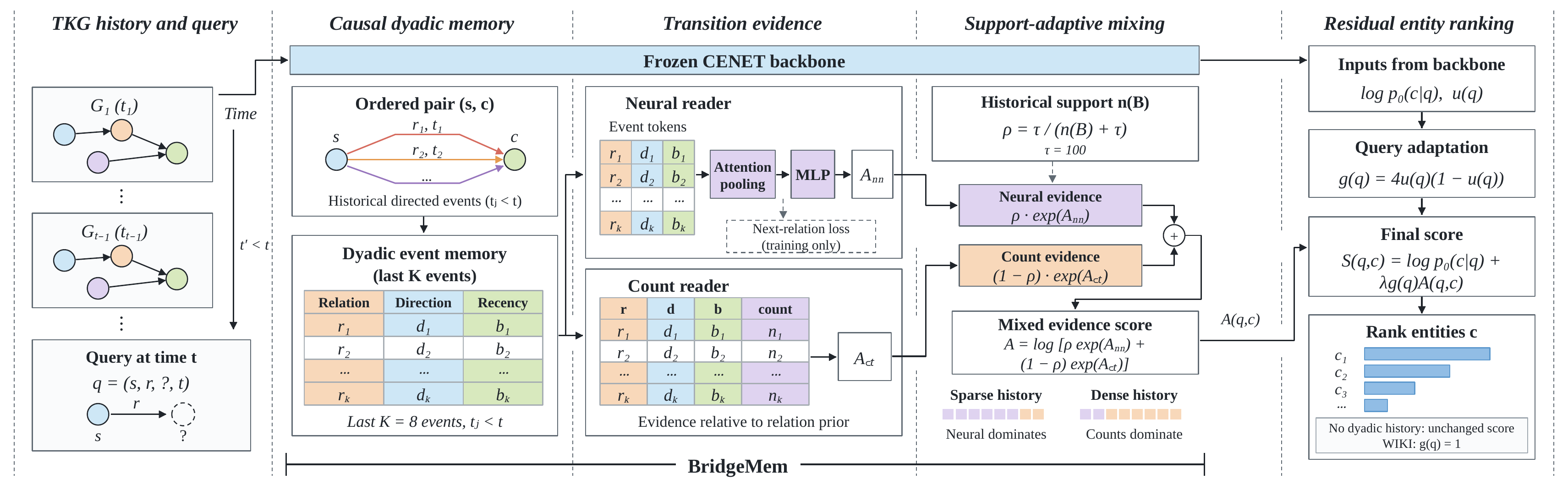}%
}
\caption{Overview of \method{}. A frozen full-vocabulary forecaster supplies
base scores and oracle uncertainty. For each candidate, \method{} retrieves
strictly prior events on the ordered actor--candidate dyad, combines neural and
count-based transition evidence according to its support, and applies the
resulting likelihood-ratio residual only when the backbone is uncertain.}
\label{fig:method}
\end{figure*}

\subsection{Causal Dyadic State}

Let $\mathcal G_{<t}$ denote all facts strictly before $t$. We build on a frozen CENET forecaster, which for an object query $q=(a,r,?,t)$ returns full-vocabulary scores $p_0(c\mid q)$ over candidates $c\in\mathcal E$ together with a query-level history-oracle probability $u_q$. Subject queries are handled by reciprocal relations in the same form. The residual $A(q,c)$ is built from evidence specific to the ordered pair $(a,c)$. The dyadic state $B_{a,c,t}$ contains the $K=8$ most recent events between $a$ and $c$ within a window of 65 snapshots. Event $j$ records its original relation $r_j$, its orientation $d_j\in\{0,1\}$ relative to $a$, and its lag $\Delta t_j=t-t_j$. Direction is kept because an edge $a\to c$ and its reverse can be followed by different transitions even when they carry the same relation label. A strict search boundary excludes all events at $t$, so both training and evaluation use exactly the information available immediately before the forecast. We use logarithmic recency bins with edges $\{0,1,4,16,64\}$ and form
\begin{equation}
\begin{aligned}
 \mathbf z_j&=\mathbf e_{r_j}+\mathbf e_{d_j}+\mathbf e_{b(\Delta t_j)},\\
 \alpha_j&=\operatorname{softmax}_j(\mathbf w^\top\tanh\mathbf z_j),\qquad
 \mathbf m=\sum_{j\in B_{a,c,t}}\alpha_j\mathbf z_j.
\end{aligned}
 \label{eq:memory}
\end{equation}
The sparse index is keyed by the ordered actor--candidate pair, and storing both orientations once lets the same index serve object queries and reciprocal subject queries. Attention weighting allows a recent salient event to dominate the pooled vector, while older transitions are still retained. Truncation bounds every read by $K$. Candidates with no qualifying event receive zero residual and keep their backbone score, so \method{} changes only rank comparisons for which dyadic evidence exists.

\subsection{Neural and Count-Based Transition Evidence}

The attention state predicts the next relation on the dyad,
\begin{equation}
\begin{aligned}
 p_\theta(r\mid B)&=\operatorname{softmax}
 \left(\operatorname{MLP}(\operatorname{LN}(\mathbf m))+\log\boldsymbol\pi\right)_r,\\
 A_{\mathrm{nn}}&=\log p_\theta(r\mid B)-\log\pi_r,
\end{aligned}
 \label{eq:neural}
\end{equation}
where $\boldsymbol\pi$ is the marginal relation prior. Subtracting $\log\pi_r$ measures how much the retrieved history raises the odds of $r$ relative to its corpus frequency, so a globally frequent relation is not rewarded for frequency alone. The neural reader pools several events and shares parameters across relations and recency bins. It may, however, smooth away a repeated transition that is already well estimated by data. In parallel, we retain such high-support evidence through smoothed transition probabilities $p_{\mathrm{ct}}(r\mid d_j,r_j,b_j)$ estimated from the causal training prefix. Evidence from the retrieved events is accumulated as
\begin{equation}
 A_{\mathrm{ct}}=\log\!\sum_{j\in B_{a,c,t}}
 p_{\mathrm{ct}}(r\mid d_j,r_j,b_j)-\log\pi_r.
 \label{eq:count}
\end{equation}
Because the count reader conditions on relation, direction, and recency rather than entity identities, it can reuse transitions observed on other dyads while still preserving the local event that triggered the lookup. Exact counts are reliable for frequent transition contexts, but their variance is high on rare ones. The neural reader behaves differently. It shares strength across sparse observations, at the cost of blurring sharp empirical regularities. Let $n_B$ be the total count supporting the retrieved transition tokens and $\rho_B=\tau/(n_B+\tau)$. We combine the two likelihood ratios in probability space,
\begin{equation}
 A(q,c)=\log\left[(1-\rho_B)e^{A_{\mathrm{ct}}}+\rho_B e^{A_{\mathrm{nn}}}\right],
 \qquad \tau=100.
 \label{eq:shrinkage}
\end{equation}
High-support dyads therefore keep the empirical estimator. Sparse dyads back off smoothly to the learned one. Because the mixture is taken before the final logarithm, $A(q,c)$ remains a likelihood-ratio correction that can be added directly to the backbone log score.

\subsection{Uncertainty-Gated Residual}

CENET's oracle decides whether historical or non-historical entities should be favored. A correction is most useful when this decision is uncertain, so we use $g(q)=4u_q(1-u_q)$ and score
\begin{equation}
 S(q,c)=\log\max\{p_0(c\mid q),\epsilon\}+\lambda g(q)A(q,c),
 \label{eq:score}
\end{equation}
where $\epsilon=10^{-12}$. The gate vanishes at confident oracle decisions and peaks at $u_q=0.5$. It is deterministic and reuses the uncertainty the frozen model already produces, adding no trainable parameters. The residual changes candidate scores only when the backbone is uncertain about the query and the candidate has a nonempty dyadic history. This prevents a transition match from overriding a confident full-vocabulary prediction. For WIKI, whose released CENET oracle is effectively discrete, validation selects the ungated form $g(q)=1$. The coefficient $\lambda$ is selected on a fixed validation grid and equals 1 on ICEWS 2014, ICEWS 2018, and GDELT, 2 on ICEWS 2005--2015, and 10 on WIKI. The backbone is kept fixed, isolating the effect of the residual.

\subsection{Training and Selection}

Each training example asks the encoder to predict the relation of the next event on a dyad from strictly earlier events on that dyad. We minimize cross-entropy over the $2|\mathcal R|$ directed target relations. The objective is self-supervised and never requires scoring the full entity vocabulary during training. The count reader is estimated from the same causal prefix but has no learned parameters. Only the event embeddings, attention vector, and transition head receive gradients.

A chronological 90/10 split inside the official training prefix selects the epoch by validation negative log likelihood, and training stops after three non-improving checks. We then refit for the selected budget on the complete training prefix. Aggregate validation MRR selects $\lambda$, the gate, and the shrinkage setting under the constraint that none of MRR or Hits@$\{1,3,10\}$ decreases relative to the frozen backbone. After selection, the encoder and count table are refitted on train plus validation and the test split is evaluated once. With the backbone frozen, training cost scales with the number of sampled dyadic transitions, while inference requires at most \(K\) indexed reads per candidate.

\begin{table*}[!t]
\caption{Complete filtered entity-forecasting matrix. Each dataset
repeats MRR, H@1, H@3, and H@10. The last block reports \method{} and its
per-metric improvement over the strongest prior method; all deltas are positive.}
\label{tab:full-metrics}
\centering

\begingroup
\ninept
\renewcommand{\arraystretch}{0.88}
\setlength{\tabcolsep}{0pt}

\begin{tabular*}{\textwidth}{
@{\extracolsep{\fill}}
l*{20}{r}
@{}
}
\toprule

Method
& \multicolumn{4}{c}{ICEWS 2014}
& \multicolumn{4}{c}{ICEWS 2018}
& \multicolumn{4}{c}{ICEWS 05--15}
& \multicolumn{4}{c}{GDELT}
& \multicolumn{4}{c}{WIKI} \\

\cmidrule(lr){2-5}
\cmidrule(lr){6-9}
\cmidrule(lr){10-13}
\cmidrule(lr){14-17}
\cmidrule(lr){18-21}

& MRR & H@1 & H@3 & H@10
& MRR & H@1 & H@3 & H@10
& MRR & H@1 & H@3 & H@10
& MRR & H@1 & H@3 & H@10
& MRR & H@1 & H@3 & H@10 \\
\midrule

CyGNet (2021)
& .3863 & .2871 & .4324 & .5815
& .2767 & .1799 & .3145 & .4702
& .4056 & .2993 & .4576 & .6116
& .2056 & .1280 & .2198 & .3583
& .6580 & .5706 & .7124 & .8251 \\

RE--GCN (2021)
& .4221 & .3185 & .4718 & .6215
& .3249 & .2235 & .3664 & .5238
& .4670 & .3618 & .5236 & .6691
& .1995 & .1265 & .2124 & .3422
& .7763 & .7386 & .8034 & .8368 \\

TiRGN (2022)
& .4478 & .3411 & .5062 & .6493
& .3366 & .2311 & .3811 & .5434
& .4950 & .3861 & .5559 & .7016
& .2173 & .1364 & .2341 & .3778
& .8164 & .7780 & .8509 & .8710 \\

CENET (2023)
& .5683 & .5206 & .5857 & .6617
& .5253 & .4831 & .5384 & .6055
& .6914 & .6560 & .7038 & .7595
& .6428 & .6161 & .6491 & .6899
& .8681 & .8660 & .8702 & .8710 \\

HRI (2024)
& .3760 & .2970 & .4154 & .5242
& .2839 & .2035 & .3203 & .4378
& .4434 & .3504 & .4963 & .6170
& .2465 & .1667 & .2714 & .4001
& .8155 & .7736 & .8571 & .8701 \\

TPAR (2024)
& .3369 & .2492 & .3686 & .5073
& .2312 & .1456 & .2581 & .4004
& .3387 & .2460 & .3767 & .5223
& .1583 & .0993 & .1620 & .2717
& .4977 & .4769 & .5106 & .5275 \\

CSI (2024)
& .3306 & .2423 & .3654 & .5002
& .2481 & .1593 & .2786 & .4239
& .3832 & .2882 & .4275 & .5638
& .1682 & .1059 & .1760 & .2866
& .5173 & .4931 & .5310 & .5568 \\

INFER (2025)
& .3054 & .2209 & .3350 & .4718
& .2369 & .1523 & .2636 & .4056
& .3547 & .2587 & .3964 & .5443
& .1661 & .1046 & .1723 & .2845
& .5051 & .4846 & .5172 & .5375 \\

EST (2026)
& .2504 & .1589 & .2843 & .4343
& .1990 & .1186 & .2182 & .3645
& .3125 & .2155 & .3525 & .5064
& .1535 & .0956 & .1566 & .2647
& .4993 & .4703 & .5155 & .5513 \\

\midrule

\textbf{\method{}}
& \textbf{.5896} & \textbf{.5429} & \textbf{.6073} & \textbf{.6791}
& \textbf{.5417} & \textbf{.4995} & \textbf{.5549} & \textbf{.6218}
& \textbf{.7130} & \textbf{.6788} & \textbf{.7271} & \textbf{.7782}
& \textbf{.6540} & \textbf{.6282} & \textbf{.6599} & \textbf{.7002}
& \textbf{.8709} & \textbf{.8707} & \textbf{.8711} & \textbf{.8712} \\

\textit{Gain vs.\ best (all +)}
& .0213 & .0223 & .0216 & .0174
& .0164 & .0164 & .0165 & .0163
& .0216 & .0228 & .0233 & .0187
& .0112 & .0121 & .0108 & .0103
& .0028 & .0047 & .0009 & .0002 \\

\bottomrule
\end{tabular*}

\endgroup
\end{table*}

\section{Experiments}

\subsection{Experimental Setup}

We use ICEWS 2014, ICEWS 2018, and ICEWS 2005--2015 with the chronological splits established for temporal KG evaluation \cite{garciaduran2018sequence}. GDELT is derived from the global event database \cite{leetaru2013gdelt}, while WIKI follows the temporal validity benchmark \cite{leblay2018validity}. The five datasets contain 7.1k, 23.0k, 10.5k, 7.7k, and 12.6k entities, with 230, 256, 251, 240, and 24 relations, respectively. Every system predicts objects and reciprocal subjects, and reports query-micro filtered MRR and Hits@$\{1,3,10\}$. All benchmark ranks are computed on the complete query sets. Filtering follows the released CENET protocol, removing non-gold entities recorded as another true answer to the same directed query in the evaluated data.

We compare CyGNet, RE-GCN, TiRGN, CENET, HRI, TPAR, CSI, INFER, and EST. Table~\ref{tab:full-metrics} lists their publication years. We rerun all baselines on the same processed files using the shared evaluation protocol. To isolate the effect of the residual, we keep the data split, reciprocal-query construction, filtering code, and metric aggregation identical across methods. RE-GCN and TiRGN run for at most 500 epochs, are validated after every epoch, and stop after five consecutive validation-MRR checks without a strict improvement; only the validation-best checkpoint is read once on test. CENET uses its fixed 30-epoch backbone plus 20-epoch oracle schedule, while CyGNet uses a 30-epoch cap with patience of three. The matched backbone is the released CENET model and checkpoint for each dataset.

\method{} uses 64-dimensional event embeddings, a 128-unit transition head, AdamW at $10^{-3}$, batches of 8,192 transition examples, and at most 20 selection epochs. Each run samples at most two million transition examples and uses patience of three for internal chronological validation. With $|\mathcal R|$ original relations, the memory adds $322|\mathcal R|+8{,}961$ trainable parameters, while the empirical transition table has no learned parameters. Each run uses one of five fixed seeds (42--46), which determine initialization, negative sampling, and data-loader order; reported results are averaged across the five runs.

We retain complete per-query ranks within every run for aggregate and mechanism analyses. Selection is locked before test evaluation. Mean validation MRR improves on all five datasets, and no dataset shows a decrease in mean validation H@1, H@3, or H@10. The per-dataset validation MRR before and after selection is 0.5863 /0.5433 /0.6919 /0.6455 /0.8523 and 0.6003 /0.5556 /0.7098 /0.6551 /0.8536 on ICEWS 2014, ICEWS 2018, ICEWS 2005--2015, GDELT, and WIKI, respectively.

\subsection{Main Results}

Table~\ref{tab:full-metrics} reports all five benchmarks and all four metrics. \method{} improves MRR over the strongest prior result by 0.0213 on ICEWS 2014, 0.0164 on ICEWS 2018, 0.0216 on ICEWS 2005--2015, 0.0112 on GDELT, and 0.0028 on WIKI. Every H@1, H@3, and H@10 entry improves as well, giving 20 strict wins in 20 comparisons. The smallest margin is WIKI H@10 at 0.0002 over TiRGN; the largest is ICEWS 2005--2015 H@3 at 0.0233 over CENET.

The gains are concentrated in the top-ranked candidates rather than in low-ranked entities. On the three ICEWS datasets, the improvements over the strongest prior method are 0.0164--0.0228 at H@1 and 0.0165--0.0233 at H@3, comparable to or larger than the corresponding H@10 gains. GDELT shows the same pattern at a larger scale, where the residual improves H@1/H@3/H@10 by 0.0121/0.0108/0.0103 over more than 610k bidirectional test queries. WIKI is already near saturation under CENET, yet all four metrics still improve, with most of the remaining gain at H@1. This indicates that the residual reorders near-tied top candidates rather than merely promoting low-ranked entities. The final row of Table~\ref{tab:full-metrics} reports the signed gain over the best of the nine prior methods separately for every metric.

\subsection{Mechanism Control and Efficiency}

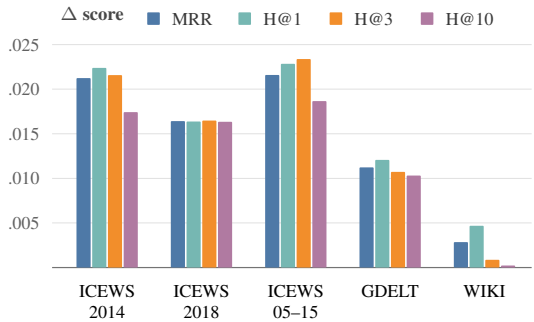
\begin{figure}[!t]
\centering
\definecolor{bmBlue}{HTML}{4E79A7}
\definecolor{bmTeal}{HTML}{76B7B2}
\definecolor{bmOrange}{HTML}{F28E2B}
\definecolor{bmPurple}{HTML}{B07AA1}
\begin{tikzpicture}[x=1.04cm,y=118cm,font=\fontsize{6.8}{7.2}\selectfont]
\draw[black!12,line width=.25pt] (.15,0.005) -- (6.25,0.005);
\node[anchor=east,text=black!62] at (.10,0.005) {.005};
\draw[black!12,line width=.25pt] (.15,0.010) -- (6.25,0.010);
\node[anchor=east,text=black!62] at (.10,0.010) {.010};
\draw[black!12,line width=.25pt] (.15,0.015) -- (6.25,0.015);
\node[anchor=east,text=black!62] at (.10,0.015) {.015};
\draw[black!12,line width=.25pt] (.15,0.020) -- (6.25,0.020);
\node[anchor=east,text=black!62] at (.10,0.020) {.020};
\draw[black!12,line width=.25pt] (.15,0.025) -- (6.25,0.025);
\node[anchor=east,text=black!62] at (.10,0.025) {.025};
\draw[black!55,line width=.45pt] (.15,0) -- (6.25,0);
\node[anchor=south west,text=black!72,font=\scriptsize\bfseries] at (.15,.0268) {$\Delta$ score};
\fill[bmBlue,rounded corners=.35pt] (0.46,0) rectangle +(.18,0.021237);
\fill[bmTeal,rounded corners=.35pt] (0.66,0) rectangle +(.18,0.022385);
\fill[bmOrange,rounded corners=.35pt] (0.86,0) rectangle +(.18,0.021571);
\fill[bmPurple,rounded corners=.35pt] (1.06,0) rectangle +(.18,0.017433);
\fill[bmBlue,rounded corners=.35pt] (1.66,0) rectangle +(.18,0.016414);
\fill[bmTeal,rounded corners=.35pt] (1.86,0) rectangle +(.18,0.016379);
\fill[bmOrange,rounded corners=.35pt] (2.06,0) rectangle +(.18,0.016480);
\fill[bmPurple,rounded corners=.35pt] (2.26,0) rectangle +(.18,0.016349);
\fill[bmBlue,rounded corners=.35pt] (2.86,0) rectangle +(.18,0.021586);
\fill[bmTeal,rounded corners=.35pt] (3.06,0) rectangle +(.18,0.022845);
\fill[bmOrange,rounded corners=.35pt] (3.26,0) rectangle +(.18,0.023376);
\fill[bmPurple,rounded corners=.35pt] (3.46,0) rectangle +(.18,0.018664);
\fill[bmBlue,rounded corners=.35pt] (4.06,0) rectangle +(.18,0.011228);
\fill[bmTeal,rounded corners=.35pt] (4.26,0) rectangle +(.18,0.012077);
\fill[bmOrange,rounded corners=.35pt] (4.46,0) rectangle +(.18,0.010733);
\fill[bmPurple,rounded corners=.35pt] (4.66,0) rectangle +(.18,0.010312);
\fill[bmBlue,rounded corners=.35pt] (5.26,0) rectangle +(.18,0.002853);
\fill[bmTeal,rounded corners=.35pt] (5.46,0) rectangle +(.18,0.004690);
\fill[bmOrange,rounded corners=.35pt] (5.66,0) rectangle +(.18,0.000871);
\fill[bmPurple,rounded corners=.35pt] (5.86,0) rectangle +(.18,0.000230);
\node[anchor=north,align=center] at (0.85,-.0010) {\shortstack{ICEWS\\2014}};
\node[anchor=north,align=center] at (2.05,-.0010) {\shortstack{ICEWS\\2018}};
\node[anchor=north,align=center] at (3.25,-.0010) {\shortstack{ICEWS\\05--15}};
\node[anchor=north,align=center] at (4.45,-.0010) {GDELT};
\node[anchor=north,align=center] at (5.65,-.0010) {WIKI};
\fill[bmBlue,rounded corners=.35pt] (1.36,.0272) rectangle +(.16,.0014);
\node[anchor=west,text=black!78] at (1.57,.0279) {MRR};
\fill[bmTeal,rounded corners=.35pt] (2.52,.0272) rectangle +(.16,.0014);
\node[anchor=west,text=black!78] at (2.73,.0279) {H@1};
\fill[bmOrange,rounded corners=.35pt] (3.68,.0272) rectangle +(.16,.0014);
\node[anchor=west,text=black!78] at (3.89,.0279) {H@3};
\fill[bmPurple,rounded corners=.35pt] (4.84,.0272) rectangle +(.16,.0014);
\node[anchor=west,text=black!78] at (5.05,.0279) {H@10};
\end{tikzpicture}
\caption{Test-set gain from the complete count/neural transition residual over
the identical frozen CENET scores. All 20 metric deltas are positive; full
metric values appear in Table~\ref{tab:full-metrics}.}
\label{fig:ablation-gains}
\end{figure}

\begin{table}[!t]
\caption{Current \method{} fitting cost on one NVIDIA H20. All times are in
seconds. Added parameters exclude the frozen CENET backbone. Select+fit
includes chronological epoch selection and the train-only refit; Final refit
trains on train plus validation for the selected epoch budget.}
\label{tab:efficiency}
\centering

\begingroup
\ninept
\renewcommand{\arraystretch}{0.92}
\setlength{\tabcolsep}{0pt}

\begin{tabular*}{\columnwidth}{
@{\extracolsep{\fill}}
lrrrrr
@{}
}
\toprule
Dataset & Params. & Epoch & Trans. & Select+fit & Final refit \\
\midrule
ICEWS 2014  & 83,021 & 20 & 88,988    & 86.2   & 55.0  \\
ICEWS 2018  & 91,393 & 20 & 448,044   & 329.7  & 194.3 \\
ICEWS 05--15 & 89,783 & 20 & 476,338  & 350.4  & 217.6 \\
GDELT       & 86,241 & 18 & 2,000,000 & 1424.2 & 656.7 \\
WIKI        & 16,689 & 15 & 1,048,280 & 613.3  & 340.6 \\
\bottomrule
\end{tabular*}

\endgroup
\end{table}

Figure~\ref{fig:ablation-gains} reports the effect of removing the learned and count-based transition residual while keeping the backbone, candidate set, filter, and evaluated queries fixed. The resulting metric drops equal the gains reported above, and all 20 metric deltas are positive. The effect is strongest on ICEWS 2005--2015 and ICEWS 2014, where multi-year actor histories provide more cross-relation transitions. It remains positive on dense GDELT, so the memory is not restricted to a small benchmark. The smaller WIKI gain is also informative. With only 24 relations and a backbone MRR of 0.8681, little ranking error remains, yet dyadic evidence still improves H@1 by 0.0047 without reducing the broader cutoffs.

Table~\ref{tab:efficiency} shows that the learned memory adds only 16.7k--91.4k parameters, far below the cost of retraining the frozen entity forecaster. Runtime grows with the number of sampled transitions, not with the entity vocabulary. Even GDELT, capped at two million transition examples, requires 1,424 seconds for selection plus the train-only refit and 657 seconds for the final train-plus-validation refit, while the other final refits finish in under six minutes. The improvement therefore comes from what is read at scoring time. It does not come from increased backbone capacity.

\section{Conclusion}

\method{} brings dyadic transition evidence to temporal knowledge graph forecasting. It retrieves the events on the ordered actor--candidate pair that strictly precede the forecast, encodes their relations, directions, and lags, and converts them into a likelihood-ratio residual on top of a frozen CENET forecaster. Count-based and neural estimates are mixed according to transition support. The backbone's history-oracle uncertainty then gates the correction. The backbone forecaster is never retrained, and only a small number of additional parameters are introduced, so the residual can be attached to an existing full-vocabulary scorer without changing its training procedure or its inference budget. The residual is therefore complementary to, rather than a replacement for, the backbone's entity-level scoring. On five benchmarks, the residual improves all 20 reported MRR and Hits@$\{1,3,10\}$ comparisons over the strongest of nine prior methods, while adding only 16.7k--91.4k parameters. The gains are largest on the datasets with rich multi-year interaction histories, where the dyadic state has more cross-relation transitions to draw on, and they remain positive on WIKI even though the backbone is already close to saturation. These results suggest that the dyadic transition signal is complementary to what entity-level and path-level forecasters already capture.

Our method has several limitations. The dyadic state uses a fixed window of $K=8$ events and a fixed support threshold $\tau=100$ across all datasets. The gate and $\lambda$ are selected on validation, not learned, and we did not tune these per dataset. Future work can learn the window, the mixing weight, and the gate jointly with the encoder, and can attach the same residual to recurrent or path-based forecasters to test whether the gains transfer across backbones.

\vfill\pagebreak

\bibliographystyle{IEEEbib}
\bibliography{refs}

\end{document}